# Hyperspectral Images Classification Based on Multi-scale Residual Network

Xiangdong Zhang, Tengjun Wang, Yun Yang

*School of Geology Engineering and Geomatics, Chang'an University, Xi'an, Shaanxi 710054, China*

**Abstract**   Because hyperspectral remote sensing images contain a lot of redundant information and the data structure is highly non-linear, leading to low classification accuracy of traditional machine learning methods. The latest research shows that hyperspectral image classification based on deep convolutional neural network has high accuracy. However, when a small amount of data is used for training, the classification accuracy of deep learning methods is greatly reduced. In order to solve the problem of low classification accuracy of existing algorithms on small samples of hyperspectral images, a multi-scale residual network is proposed. The multi-scale extraction and fusion of spatial and spectral features is realized by adding a branch structure into the residual block and using convolution kernels of different sizes in the branch. The spatial and spectral information contained in hyperspectral images are fully utilized to improve the classification accuracy. In addition, in order to improve the speed and prevent overfitting, the model uses dynamic learning rate, BN and Dropout strategies. The experimental results show that the overall classification accuracy of this method is 99.07% and 99.96% respectively in the data set of Indian Pines and Pavia University, which is better than other algorithms.
**OCIS codes** 280**.**4991**;** 100**.**2960**;** 100**.**4145

## 1. Introduction

Hyperspectral remote sensing is a new type of remote sensing technology that emerged in the early 1980s. It can obtain spectral information of hundreds of continuous wave bands of surface objects. While recording spectral information of ground features, it also records spatial information, achieving "map and spectrum integration." Hyperspectral images have nanometer-level spectral resolution, which can reflect the subtle differences in spectral dimensions of different features, greatly improving the ability to distinguish and identify features. As the technology matures, hyperspectral images are widely used in the fields of fine agriculture [1], environmental monitoring, etc. [2-3], and classification is a key research link in many fields. The purpose of hyperspectral image classification is to give each pixel a unique category label, thereby completing the automatic recognition of features. However, due to the high-dimensional characteristics of hyperspectral data and the lack of label samples, it faces huge challenges in the field of classification.

In the early stages of studying hyperspectral image classification, common machine learning methods such as support vector machine (SVM) [4], polynomial logistic regression (MLR) [5], and extreme learning machine (ELM) [6] only use spectral features as classification. The basis of the method is that no spatial information is used. Due to the limited number of label samples, Hughes phenomenon that classification accuracy decreases with the increase of feature dimension will occur in most methods. Among them, SVM uses kernel transform technology to obtain a hyperplane from a small number of samples, which can classify high-dimensional data, and obtain good classification results on small sample data sets. In order to further improve the classification accuracy, some scholars continue to introduce spatial information in the machine learning method based on spectral features. For example, Kang et al. [7] used edge retention filtering technology combined with spatial spectral information to improve the classification accuracy of SVM; Li et al. [8] A method of combining Markov random fields and subspace polynomial logistic regression is proposed to classify hyperspectral images in the spectral and spatial domains; Fang et al. [9] consider that hyperspectral pixels can usually be expressed as the same type of ordinary The linear combination of pixels merges the spatial information in the neighborhood of each pixel into a sparse representation model. Although the above method based on spatial spectrum combined features achieves better classification results than ordinary methods, it relies heavily on manual design of classification features, and the artificially extracted shallow features are not enough to distinguish the subtle differences between different features and similar features Large differences between [10].

In recent years, as deep learning has made major breakthroughs in the field of computer vision such as image classification and target detection, researchers have proposed a large number of hyperspectral image classification algorithms based on deep learning. In 2014, Chen et al. [11] first applied stacked self-encoding machines (SAE) to hyperspectral image classification problems, and built a deep network by stacking multiple layers of autoencoders to improve classification accuracy; in 2015, Hu et al. [12] For the first time, CNN was introduced into hyperspectral classification, but only spectral information was used. In order to use both spectral and spatial information in CNN, Chen et al. [13] proposed a 3D-CNN-based deep feature extraction method, which provided new research ideas for the application of CNN in the field of hyperspectral image classification. In order to make full use of the rich spatial spectrum information in hyperspectral images, Zhong et al. [14] proposed the SSRN model, which used continuous residual units to learn spectral and spatial features, and constructed a deep network structure to achieve different levels of features Compared with the above method, the best classification results are obtained. However, when the hyperspectral data involved in training in the deep learning method is reduced, the classification accuracy has declined to varying degrees. Considering the lack of sample labels in actual research, how to achieve accurate classification under the condition of small sample hyperspectral data is a challenging problem.

In order to solve this problem, this paper proposes a 3D-CNN-based multi-scale residual network model for extracting multi-scale spectral features and spatial features of hyperspectral images. On the one hand, the model uses residual learning to alleviate the degradation problem of deep networks; on the other hand, by introducing a branch structure containing convolution kernels of different sizes [15], the extraction and fusion of multi-scale spatial spectrum features are realized. The spatial spectrum information of hyperspectral data is used to improve the classification accuracy on small samples of hyperspectral data. In addition, since $1 \times 1 \times 1$ convolution is used in many places to reduce the dimensionality of the channel, the number of parameters of the model is effectively reduced, and the network performance is improved while ensuring the accuracy. The network model does not require pre-processing and post-processing, and is trained in an end-to-end manner. It has good scalability for different hyperspectral data sets.

# 2. Methods

## 2.1. Residual Learning

Compared with the shallow network, the deep network has stronger learning ability and feature expression ability. When the network reaches a certain depth, continuing to increase the number of layers will not bring performance improvement, but will cause network degradation, that is, with the network layer As the number increases, the accuracy rate on the training set gradually saturates or even decreases. In order to solve the degradation problem, He et al. [16] proposed residual learning, as shown in Figure 1. Suppose the network input is x and the desired feature map is H (x). The residual map F (x) = H (x) -x is realized by adding jump connections, then the original map can be expressed as F (x) + x. In the forward propagation process, when the shallow layer completes the feature extraction, residual learning can enable the deep network to achieve identity mapping, so that the network can increase the number of layers while avoiding degradation problems.

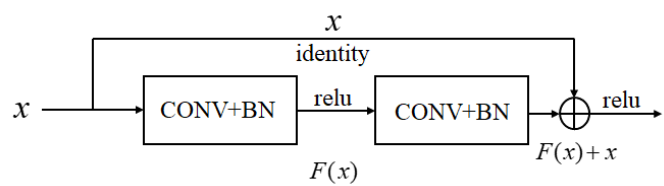

Fig. 1 Block of residual learning

## 2.2. Multi-scale Spatial-Spectral Features Extraction

Due to the wide range of hyperspectral images, high spectral resolution, and strong spatial correlation, this paper designed multi-scale feature extraction units for spectral and spatial dimensions, respectively. By adding a branch structure to the residual module, and using convolution kernels of different sizes in each branch to obtain different scale features of the input image, then use the stitching operation to connect the output feature maps, and finally pass $1 \times 1$ The $\times 1$ convolution reduces the dimension of the channel to achieve the feature fusion with the original input.

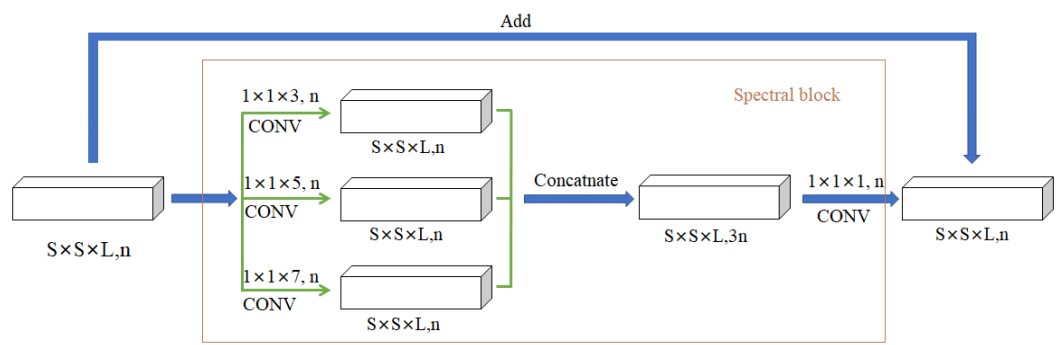

Fig. 2 Multi-scale spectral feature extraction block

As shown in Fig. 2, the multiscale spectral feature extraction unit uses convolution kernels with a size of 1×1×m (m = 3, 5, 7) respectively, and keeps the input and output sizes of the convolutional layer consistent by edge filling. The unit input is and the output is, then the spectral feature extraction unit can be expressed as:

$$x_{i+1} = x_i + H\left\{T\left\{\left[F_1(x_i), F_2(x_i), F_3(x_i)\right]\right\}\right\}$$

In the formula: $F_1(\cdot), F_2(\cdot), F_3(\cdot)$, respectively represent the nonlinear operation of convolution kernels of different spectral dimensions, which means that the feature maps output through the convolution layer are connected, and that the channel is reduced in dimension.

In the spatial feature extraction unit shown in Fig. 3, convolution kernels of size r × r × 1 (r = 1, 3, 5) are used, the input of this unit is $x_j$ and the output is $x_{j+1}$, then the spatial feature extraction process can be expressed as:

$$x_{j+1} = x_j + H\left\{T\left\{\left[D_1(x_j), D_2(x_j), D_3(x_j)\right]\right\}\right\}$$

In the formula: $D_1(\cdot)$, $D_2(\cdot)$, $D_3(\cdot)$, respectively represent the non-linear operation of convolution kernels of different spatial sizes, which means that the feature maps output through the convolution layer are connected, and that the channel is reduced in dimension.

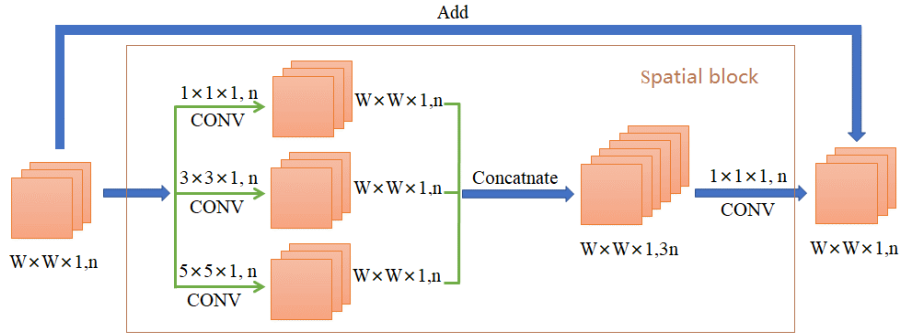

Fig. 3 Multi-scale spatial feature extraction block

## 2.3. Multi-scale Residual Network

Figure 4 shows the multi-scale residual network model (MSRN) proposed in this paper, which consists of three convolutional layers, spectral and spatial feature extraction modules, an average pooling layer and a fully connected layer. Taking the Indian Pines data set as an example, a pixel block with a size of 11 × 11 × 200 is used as the model input, the size of the convolution kernel of the first convolution layer is 1 × 1 × 7, and the number of convolution kernels is 24 Considering that there is a large amount of redundant information in the hyperspectral data, the sampling step size is set to (1, 1, 2); the output of the spectral feature extraction module is 24 feature maps with a size of 11 × 11 × 97, through two consecutive 3D convolution converts it into the input of the spatial feature extraction module; after completing the extraction of the spatial spectrum features, the average pooling operation converts all feature maps into feature vectors and generates category labels through the fully connected layer.

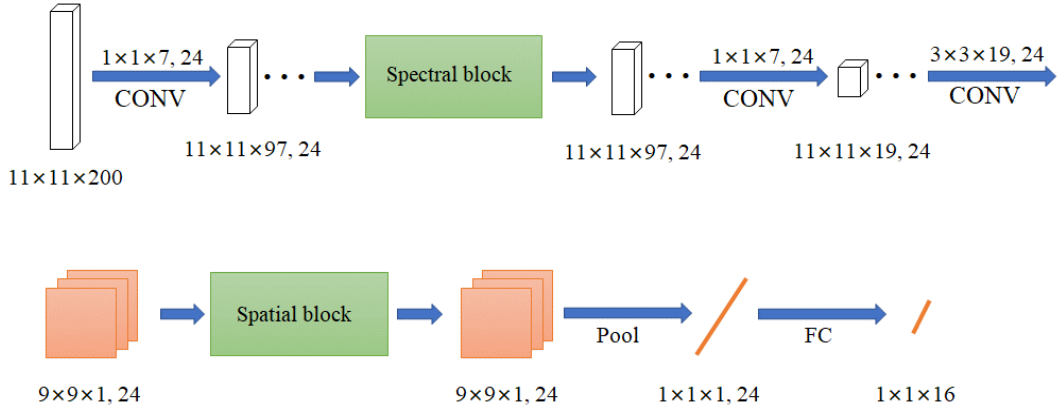

Fig. 4 Multi-scale residual network

By adding a branch structure to the residual learning module, the MSRN model widens the network while deepening the network. The researchers demonstrated in experiments that increasing the width can also improve network performance [17]. Therefore, compared with ordinary deep convolutional networks, the MSRN model proposed in this paper has better classification performance.

## 3. Experimental results and analysis

The experiments used Indian Pines (IN) and Pavia University (UP), two representative data sets, and specified the configuration of network parameters. The impact of different parameters on the model performance was analyzed. The overall accuracy (overall accuracy, OA) was used in the experiment, Average accuracy (AA) and Kappa coefficient are used as classification accuracy evaluation indicators.

### 3.1. Datasets

Indian Pines data was acquired by the Airborne Visible / Infrared Imaging Spectrometer (AVIRIS) sensor in the Indiana test site in 1996. The image size is 145 × 145 pixels and the wavelength range are 0.4 to 2.45 um. After removing 20 water vapor absorption band. The remaining 200 effective bands. The image contains a total of 16 types of features. Considering the low spatial resolution of 20m, it is easy to cause confusion between the corresponding pixels of various types of features, thus increasing the difficulty of classification. The experiment randomly selected 10%, 10%, and 80% as the training set, validation set, and test set, respectively. The corresponding sample numbers are shown in Table 1.

Table 1 Sample distribution of Indian Pines data

| Number | Class | Train | Val | Test |
| --- | --- | --- | --- | --- |
| 1 | Alfalfa | 10 | 3 | 33 |
| 2 | Corn-notill | 285 | 133 | 1011 |
| 3 | Corn-mintill | 165 | 85 | 579 |
| 4 | Corn | 48 | 24 | 165 |
| 5 | Grass-pasture | 97 | 44 | 342 |
| 6 | Grass-trees | 146 | 71 | 513 |
| 7 | Grass-pasture-mowed | 6 | 5 | 17 |
| 8 | Hay-windrowed | 96 | 57 | 325 |
| 9 | Oats | 4 | 6 | 10 |
| 10 | Soybean-nottill | 194 | 96 | 682 |
| 11 | Soybean-mintill | 490 | 265 | 1699 |
| 12 | Soybean-clean | 119 | 58 | 417 |
| 13 | Wheat | 41 | 28 | 136 |
| 14 | Woods | 252 | 138 | 875 |
| 15 | Buildings-Grass-Trees | 78 | 36 | 272 |
| 16 | Stone-Steel-Towers | 19 | 7 | 67 |
|  | TOTAL | 2050 | 1056 | 7143 |

The Pavia University data was acquired by the German Reflection Optical Spectral Imager (ROSIS) in Italy in 2003. The image size is 610 × 340, the wavelength range is 0.43 ~ 0.86μm, and the remaining 103 are available after excluding 12 bands affected by noise Band. Since the image contains only 9 types of features, and has a spatial resolution of 1.3 meters, the classification difficulty is lower than that of the IN data set. Therefore, for this data set, 5%, 5%, and 90% were randomly selected as the training set, validation set, and For the test set, the corresponding sample size is shown in Table 2.

Table 2 Sample distribution of Pavia University data

| Number | Class | Train | Val | Test |
| --- | --- | --- | --- | --- |
| 1 | Asphalt | 331 | 332 | 5968 |
| 2 | Meadows | 932 | 933 | 16784 |
| 3 | Gravels | 105 | 105 | 1889 |
| 4 | Trees | 153 | 153 | 2758 |
| 5 | Painted-Metal-Sheets | 67 | 67 | 1211 |
| 6 | Bare-Soil | 251 | 252 | 4526 |
| 7 | Bitumen | 66 | 67 | 1197 |
| 8 | Self-Blocking-Bricks | 185 | 183 | 3314 |
| 9 | Shadows | 48 | 47 | 852 |
|  | TOTAL | 2138 | 2139 | 38499 |

## 3.2. Parameters Setting

The performance of the deep learning model depends on the design of the network structure and the selection of network parameters. After the construction of the MSRN framework is completed, three parameters that affect the network training process and classification performance are analyzed experimentally, namely: learning rate, convolution kernel Quantity, enter the pixel block size. In order to evaluate the classification effect of the model on the small sample data set, the proportion of the randomly selected training set in the IN and UP data sets is small, so the batch size is set to 16, and the loss function is optimized using the RMSProp optimizer. During training, the model with the highest classification accuracy on the verification set is saved, and the test set is evaluated through the optimal model.

The learning rate is an important hyperparameter of the deep network. The proper value can make the loss function converge to the minimum value at an appropriate speed. The experiment uses the grid search technology to determine the best learning rate. The learning rate to be searched includes 0.003, 0.001, 0.0003, 0.0001, and 0.00003, a complete training and test is performed for each value, and the results show that the optimal learning rate for both IN and UP data sets is 0.0003. In addition, the learning rate decay strategy is also used in the training process. When the verification loss exceeds 5 epochs, the learning rate is reduced to half of the original one, which accelerates the convergence of the network and effectively increases the training speed.

Secondly, the number of convolution kernels determines the number of feature maps obtained by the input pixel block after convolution calculation. Too few convolution kernels will result in insufficient feature extraction, while too many convolution kernels will bring too much parameter. The network structure proposed in this paper has the same convolution kernel in each convolutional layer. The network with convolution kernel numbers of 8, 16, 24, 32, 40 is used for training and testing, as shown in Figure 5. The model with a core number of 24 achieved the highest accuracy in both IN and UP datasets.

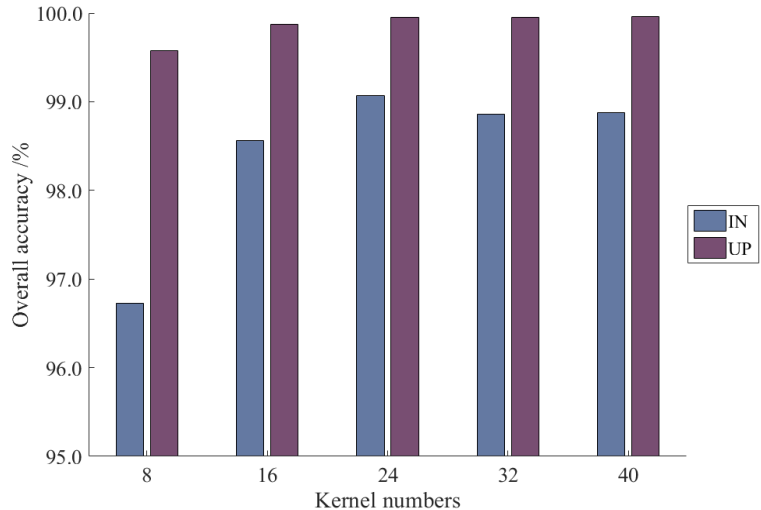

Fig. 5 OA (%) of models with different kernel numbers

In hyperspectral images, each pixel to be classified has a high probability of belonging to the same category as the pixels in its neighborhood. However, smaller neighborhoods are likely to cause insufficient receptive fields, and larger neighborhoods will cause noise problems. The choice of block size also has a certain influence on the classification accuracy. In order to select the appropriate pixel block size, the model was tested using 5 different-sized pixel blocks as network inputs. The corresponding classification accuracy is shown in Figure 6. The results show that the spatial size of the pixel block with the highest classification accuracy for IN and UP data is $13 \times 13$ and $11 \times 11$, respectively.

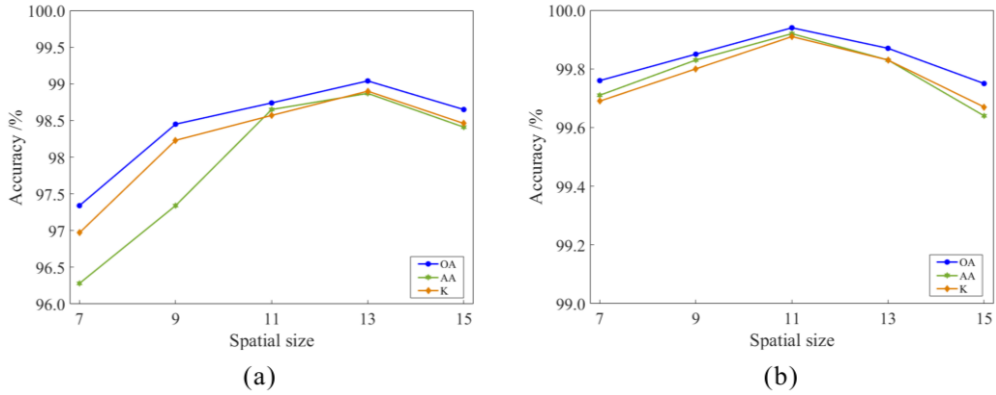

Fig. 6 Classification accuracy of different pixel block sizes. (a) IN; (b) UP

This model uses multiple methods to mitigate or avoid overfitting problems. First, choose the ReLu activation function, because the ReLu function makes the output of some neurons to 0, causing the sparseness of the network, reducing the interdependence between the parameters, thereby alleviating the problem of overfitting; second, Dropout [18] can make the network in During the training process, some neural units are randomly disabled with a certain probability, which can significantly reduce the overfitting phenomenon. In this paper, Dropout is added before the fully connected layer and its probability is set to 0.3; In addition, BN (Batch Normalization) is added after each convolutional layer. normalizes the parameters and improves the network generalization ability. Finally, an early stop strategy is added to the training process. When the verification loss exceeds 15 epochs, the training is stopped to prevent overfitting.

### 3.3. Discussion

In order to evaluate the performance of the MSRN model, it is compared with the traditional machine learning method SVM and the current mainstream deep learning methods (such as 3D-CNN and SSRN). In order to ensure the fairness of the experiment, the parameters of other comparison methods are set to the optimal, and 10% and 5% of the IN and UP data sets are randomly selected as the training set, and the same proportion as the training set is used for the test set. Table 4 lists the classification accuracy of different methods. The results show that the MSRN method proposed in this paper achieves the highest accuracy on the IN and UP data sets. For example, on an IN data set with uneven samples, training with only 10% of the data can achieve an overall accuracy of 99.09%, which is 20.25% higher than the traditional machine learning method SVM, compared with deep learning methods such as CNN and SSRN They are improved by 7.62% and 1.23% respectively, which proves the robustness of the MSRN model under the condition of small sample training data. It is precisely because the MSRN model adds a branch structure to the residual network. On the one hand, a deep network is constructed to effectively use the spectral and spatial features; on the other hand, the width of the network is increased, and the input is obtained through convolution kernels of different sizes. The information of different scales in the image is combined with the extracted multi-scale features, so MSRN obtains better classification results.

Table 4 Comparison of classification accuracy of different methods

| Methods | IN | | | UP | | |
| --- | --- | --- | --- | --- | --- | --- |
| | OA% | AA% | K×100 | OA% | AA% | K×100 |
| SVM | 78.82 | 74.66 | 76.43 | 86.22 | 85.65 | 85.96 |
| CNN | 91.45 | 89.87 | 90.36 | 96.69 | 96.20 | 95.98 |
| SSRN | 97.84 | 94.28 | 96.82 | 99.17 | 99.09 | 99.12 |
| **MSRN** | **99.07** | **98.87** | **98.90** | **99.96** | **99.94** | **99.94** |

Figures 7 - 8 show the classification results of different methods on the two data sets. It can be seen from the figure that the SVM classification map has many features that are wrongly divided. Due to the powerful feature extraction capabilities of deep learning methods, CNN Compared with the SVM, the misclassification situation has been greatly improved. The SSRN model also produces smooth classification results, but there are still noise points in certain categories. Compared with other methods, MSRN obtains the smoothest and most accurate classification map on two types of hyperspectral data sets

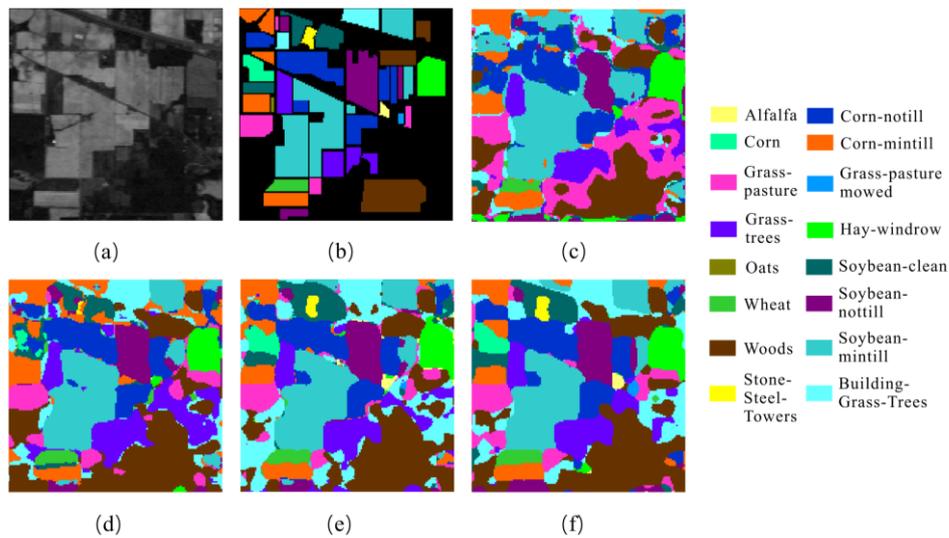

Fig. 7 Classification map of IN dataset. (a) real image in one band; (b) SVM; (c) CNN; (d) SSRN; (e) MSRN

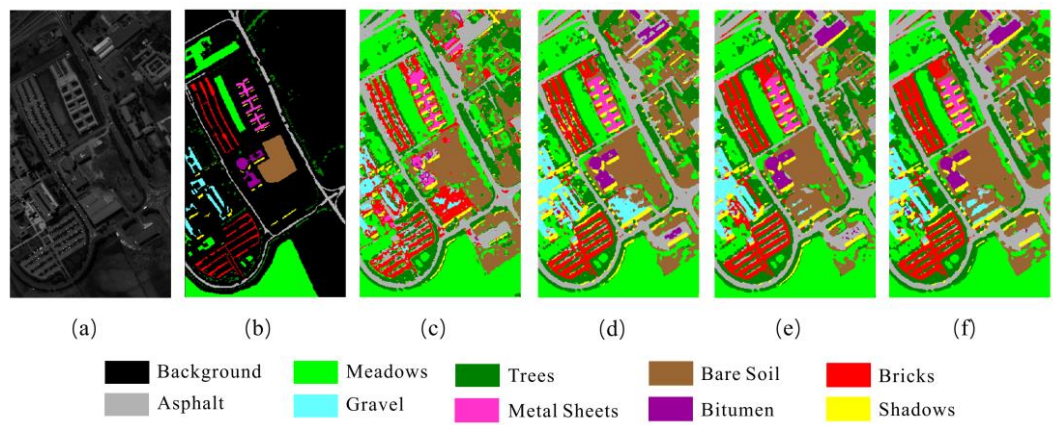

Fig. 8 Classification map of UP dataset. (a) real image; (b) SVM; (c) CNN; (d) SSRN; (e) MSRN

All the experiments in this article were completed under Windows system using Tensorflow as the back-end Keras deep learning framework. The hardware configuration is Intel Core i7-8700K, Nvidia GeForce GTX1080Ti and 64G memory. Table 5 lists the training and testing time of MSRN and other deep learning methods. Due to the use of 1×1×1 convolution in the spectral and spatial feature extraction structure to achieve channel dimensionality reduction, the network parameters are reduced, which shortens Model training test time. In addition, the introduction of early stopping strategy reduces the training time while reducing overfitting. Therefore, compared with other deep learning models, MSRN requires the least training time to achieve the best accuracy.

Table 5 Comparison of training and testing time of different algorithms

| Dataset | Time | CNN | SSRN | MS-ResNet |
| --- | --- | --- | --- | --- |
| IN | Train | 509.5 | 628.6 | 229.61 |
|  | Test | 2.42 | 3.97 | 7.46 |
| UP | Train | 1321.6 | 1034.4 | 192.96 |
|  | Test | 4.73 | 10.54 | 22.57 |

# 4. Conclusion

In order to solve the problem of low classification accuracy on small sample hyperspectral data, a deep learning framework based on multi-scale residual network was designed. The network structure introduces branch structure into the residual learning module to extract and fuse the spectral features and spatial features separately, making full use of the spatial and spectral information of hyperspectral images. The model directly uses three-dimensional cube hyperspectral data as input, without any pre-processing and post-processing, to achieve end-to-end hyperspectral image classification. The experimental results show that the model in this paper achieves 99.07% and 99.96% overall classification accuracy on the Indian Pines and Pavia University datasets, which is more computationally efficient than other methods and effectively improves the classification performance under small sample training data.

Considering the complexity of network design in deep learning methods, follow-up research will try to use AutoML technology to construct a general model suitable for hyperspectral image classification.